\documentclass{article}

    \PassOptionsToPackage{numbers, compress}{natbib}
 \usepackage[preprint]{neurips_2025}

\usepackage[utf8]{inputenc} 
\usepackage[T1]{fontenc}    
\usepackage{hyperref}       
\usepackage{url}            
\usepackage{booktabs}       
\usepackage{amsfonts}       
\usepackage{nicefrac}       
\usepackage{microtype}      
\usepackage{xcolor}         

\usepackage{graphicx}
\usepackage{amsmath}
\usepackage{multirow}
\usepackage{float}
\usepackage{pgfplots}
\pgfplotsset{compat=1.18}

\usepackage{tikz}
\usetikzlibrary{positioning,fit}
\usepackage{listings}
\usepackage{varwidth}
\usepackage{caption} 

\lstdefinestyle{sql}{
  language=SQL,
  basicstyle=\ttfamily\small,
  breaklines=true,
  columns=fullflexible
}

\tikzset{
  casebox/.style={draw, rounded corners, fill=gray!3, inner sep=6pt, align=left},
}

\newcommand{\sqlbox}[1]{%
  \begin{varwidth}{0.96\linewidth}
  \begin{lstlisting}[style=sql]
#1
  \end{lstlisting}
  \end{varwidth}
}

\title{\textsc{AgentiQL}: An Agent-Inspired Multi-Expert Framework for Text-to-SQL Generation}

\author{%
  Omid Reza Heidari\thanks{Website: \url{https://omid-reza.github.io}, 
  alternative email: \texttt{omid.orh@gmail.com}} \\
  Concordia University\\
  Montreal, Canada \\
  \texttt{o\_heidar@live.concordia.ca} \\
  \And
   Siobhan Reid \\
  Concordia University\\
  Montreal, Canada \\
  \texttt{si\_reid@live.concordia.ca} \\
  \AND
  Yassine Yaakoubi \\
  Concordia University \\
  Montreal, Canada \\
  \texttt{yassine.yaakoubi@concordia.ca} \\
}

\begin{document}

\maketitle

\begin{abstract}
LLMs have advanced text-to-SQL generation, yet monolithic architectures struggle with complex reasoning and schema diversity. We propose \textsc{AgentiQL}, an agent-inspired multi-expert framework that combines a reasoning agent for question decomposition, a coding agent for sub-query generation, and a refinement step for column selection. An adaptive router further balances efficiency and accuracy by selecting between our modular pipeline and a baseline parser. Several steps in the pipeline can be executed in parallel, making the framework scalable to larger workloads. Evaluated on the Spider benchmark, \textsc{AgentiQL} improves execution accuracy and interpretability and achieves up to 86.07\% EX with 14B models using the Planner\&Executor merging strategy. The attained performance is contingent upon the efficacy of the routing mechanism, thereby narrowing the gap to GPT-4-based SOTA (89.65\% EX) while using much smaller open-source LLMs. Beyond accuracy, \textsc{AgentiQL} enhances transparency by exposing intermediate reasoning steps, offering a robust, scalable, and interpretable approach to semantic parsing.
\end{abstract}


\section{Introduction}
\label{introduction}

Natural language to SQL (NL2SQL) technology, which enables the transformation of everyday language queries into structured SQL commands, marks a major step forward in improving data accessibility. It empowers both novice and expert users to efficiently extract meaningful insights from large and complex datasets \citep{chen2023texttosqlerrorcorrectionlanguage, liu2025surveytexttosqlerallms, tai2023exploringchainofthoughtstyleprompting, wang2021ratsqlrelationawareschemaencoding, 10.1145/3589292, li2023llmservedatabaseinterface, liu2022semanticenhancedtexttosqlparsing, wang2022protonprobingschemalinking, pourreza2024sqlgenbridgingdialectgap, pourreza2024dtssqldecomposedtexttosqlsmall, sun2024sqlpalmimprovedlargelanguage, rai2023improvinggeneralizationlanguagemodelbased}. Recent progress in large language models has considerably improved the efficacy and accuracy of NL2SQL systems, but key challenges remain. Monolithic LLM architectures often struggle with complex reasoning and with handling diverse database schemas, while static ensemble methods introduce significant computational overhead. Serving massive large language models (LLMs) in real-world environments is also costly and impractical for many applications. Moreover, most existing systems offer limited interpretability, making it difficult to understand \emph{why} a particular SQL query was generated or to identify misalignment.

To address these limitations, we propose \textsc{AgentiQL}, an agent-inspired, multi-expert architecture for NL2SQL. Instead of relying on a single monolithic LLM, \textsc{AgentiQL} decomposes query generation into specialized expert components and employs a learned router to balance accuracy and efficiency. This design yields a more interpretable, modular, and scalable framework for structured query generation. Our contributions are threefold: (i) a \textbf{Divide-and-Merge} module, where a reasoning agent decomposes natural language questions into sub-questions and a coding agent generates corresponding sub-queries that are then merged, improving interpretability through visible intermediate steps; (ii) a \textbf{Column Selection} (CS) refinement that adjusts column choices and ordering in the final SQL query, increasing alignment with user intent and boosting execution accuracy; and (iii) an \textbf{Adaptive Routing} mechanism that directs queries (e.g. XGBoost classifier\cite{Chen_2016} or a reasoning-agent "judge"), enhancing efficiency, efficacy, and robustness by adaptively allocating available resources to query complexity.

\section{Related Works}

There are three main categories of LLM-based NL2SQL approaches: \textbf{prompt engineering} \cite{dong2023c3zeroshottexttosqlchatgpt, gao2023texttosqlempoweredlargelanguage, pourreza2024chasesqlmultipathreasoningpreference, pourreza2023dinsqldecomposedincontextlearning}, supervised fine-tuning (SFT) \cite{li2024codesbuildingopensourcelanguage}, and reinforcement learning (RL)-based optimization \cite{liu2025skyrlsql}.

\textbf{Prompt engineering} has shown strong potential for NL2SQL, particularly through zero-shot approaches (e.g., using ChatGPT/GPT-4 directly \citep{dong2023c3zeroshottexttosqlchatgpt}) and few-shot chain-of-thought prompting techniques \citep{wang2025macsqlmultiagentcollaborativeframework, gao2023texttosqlempoweredlargelanguage, pourreza2023dinsqldecomposedincontextlearning}. These methods leverage pretrained LLMs with carefully crafted prompts. Nonetheless, many prompt-based methods depend on multi-path generation combined with self-consistency (majority voting) to select the optimal output, which leads to considerable inference overhead in practice.

In contrast, \textbf{SFT-based} approaches fine-tune smaller or domain-specific models on NL2SQL data to generate more controllable SQL queries (e.g., \textsc{CodeS} fine-tunes open-source LLMs for text-to-SQL \citep{li2024codesbuildingopensourcelanguage}). However, reduced parameter capacity can limit their ability to handle complex NL2SQL reasoning or generalize effectively to databases in new domains.


\section{Method}

Given a labeled dataset $\mathcal{D} = \{(x_i, s_i, y_i)\}_{i=1}^{N}$, where each natural language query $x_i$ with database schema $s_i$ is paired with the executable SQL query $y_i$, the goal is to train a model that maps $(x_i, s_i) \mapsto y_i$. We introduce \textsc{AgentiQL}, a multi-expert architecture designed to address this task through query decomposition, specialized code generation, and adaptive routing.

\subsection{Division}

\paragraph{Table Selection.} For a given question $x$ with database schema $s$, we first filter out irrelevant tables from $s$ using a reasoning LLM $f_{\text{reason}}$. Formally, this produces a reduced schema $\tilde{s} = f_{\text{reason}}(x, s)$, where $\tilde{s} \subseteq s$ contains only the tables likely needed to answer $x$, and the semantic content of the original schema remains unchanged. The reduced schema $\tilde{s}$ is passed as context to all later stages in the pipeline.

\paragraph{Question Decomposition.} Next, we employ another reasoning LLM $f_{\text{decomp}}$ to decompose the natural language query $x$ (with respect to schema $\tilde{s}$) into a set of smaller sub-questions or tasks. Formally, this step produces $\{x_1, x_2, \ldots, x_k\} = f_{\text{decomp}}(x, \tilde{s})$, where each $x_j$ is a natural language sub-question. These sub-questions are intended such that solving each one individually (on the given schema) and then merging the results will answer the original SQL query. Decomposition breaks (complex, resource-intensive) queries into one or multiple manageable natural language sub-questions and makes the reasoning process explicit.

\paragraph{Query Generation:} For each sub-question $x_j$, we use a coding LLM $f_{\text{gen}}$ (one specialized in code/SQL generation) to produce the corresponding SQL query $y_j$. This model operates in a few-shot setting, similar to the baseline. Formally, the output of this step is $f_{\text{gen}}(\{x_1, x_2, \ldots, x_k\}) = \{y_1, y_2, \ldots, y_k\}$, where each $y_i$ is the SQL query corresponding to the sub-question $x_i$. If an error is detected in a generated query, a refinement process is triggered. In this process, the same LLM is used to correct the erroneous query for up to $R$ iterations. At refinement step $r \in \{1, \ldots, R\}$, the model produces $y_i^{(r)} = f_{\text{gen}}(x_i, y_i^{(r-1)})$, correcting potential errors in the previous version.

\subsection{Merge}

The essential part of the pipeline is the ability to merge the generated sub-queries $\{y_1, y_2, \ldots, y_k\}$ into a single final SQL query $y$. Formally, this is achieved through a merge function $g$ such that $y = g(y_1, y_2, \ldots, y_k)$. We explore two strategies for $g$ in this work: \textbf{(1) Selecting the Last Sub-query:} The output SQL query is taken to be the translation of the final generated sub-question. This strategy assumes that the reasoning agent orders sub-questions such that the last one corresponds to the complete solution. It is simple and fast, but may fail if the final sub-question does not cover the entire query. \textbf{(2) Planner\&Executor:} A reasoning LLM is employed as a planner to determine how the generated sub-queries $\{y_1, \ldots, y_k\}$ should be combined. The planner produces a natural-language description or pseudocode of a merging plan, which is then executed by a coding LLM (the executor) to yield the final SQL query $y$. This approach is more general, as it does not assume that one sub-query fully answers the question, but it introduces additional computational overhead.

This divide-and-merge design naturally follows a \emph{prompt chaining workflow}, in which intermediate reasoning steps are explicit and sequential. Such chaining improves interpretability by exposing sub-questions and their corresponding sub-queries, reflecting the workflow taxonomy introduced by Anthropic for building effective agents \cite{anthropic2024effectiveagents}.

\subsection{Column Selection}
After merging, we obtain an intermediate SQL query $\hat{y}$. In a final refinement step, a reasoning LLM $f_{\text{col}}$ is given the original question $x$, schema $\tilde{s}$, and query $\hat{y}$ from the previous step. The model then performs \emph{CS}, adjusting the \texttt{SELECT} clause, ensuring output columns and their ordering precisely match the requirements of $x$. Formally, the final SQL query is obtained as $y = f_{\text{col}}(x, s, \hat{y})$ where $f_{\text{col}}$ denotes the CS function, aligning columns in $\hat{y}$ with user intent.

\subsection{Routing}

After evaluating the divide-and-merge module across multiple LLMs, we identified complementary strengths in the different approaches. According to Tables~\ref{table:percetage_of_wrong_and_correct_among_ours_and_baseline}--\ref{table:divide_and_merge_module_performance}, decomposition often captured domain-specific reasoning more effectively, whereas the baseline produced more consistent parsing in other cases. To exploit these strengths, we introduce a router that selects between the baseline or divide-and-merge pipeline based on the schema and query received. In addition to this intuition, we also tested a simple schema-level metric (the table count) as a proxy for complexity. As observed in Table~\ref{table:corr_table_count_vs_performance}, there are meaningful correlations between schema size and relative performance, suggesting that even lightweight signals can guide effective routing decisions.

The performance comparisons reported in Tables~\ref{table:impact_of_column_selection_and_merging_strategy}--\ref{table:divide_and_merge_module_performance} further confirm that, although the baselines perform well on simple queries, the divide-and-merge strategies (particularly Planner and Executor with CS) are more robust when it comes to complex reasoning tasks. These findings suggest that routing based on simple complexity measures is promising, and motivate the development of more advanced routers that combine such metrics with learned decision functions to improve efficiency and accuracy in practice.

\section{Experimental Results}

\subsection{Dataset}

Among available text-to-SQL benchmarks such as BIRD \cite{li2023llmservedatabaseinterface}, SQL-Eval \cite{defog_ai_sql_eval}, and \textsc{SkyRL-SQL} \cite{novasky_skyrl_v080_data}, the Spider \cite{yu2019spiderlargescalehumanlabeleddataset} dataset is selected for our experiments. Spider was the first large-scale benchmark proposed for the text-to-SQL task and has since been widely adopted in the literature. It contains over 10,000 natural language questions across 200+ databases with diverse schemas, covering multiple domains and complex query structures. Its widespread use ensures comparability with prior work and provides a reliable basis for assessing the effectiveness of the proposed approach.

\subsection{Baseline}

A coding LLM serves as a standard baseline for the text-to-SQL task. Specifically, we use the Qwen2.5-Coder series as the base text-to-SQL model. The model is prompted with several question-SQL examples and then directly generates the SQL for a new question in one step, using few-shot prompting, without any task-specific adaptation or additional training. We adopt a vanilla LLM as our baseline parser, which simply maps text to SQL based on the provided in-context examples.

\subsection{Metrics and Parameters}

\label{metrics_and_parameters}
Execution accuracy (EX) is adopted as the primary evaluation metric. A predicted SQL query is considered correct if it executes to the same result as the ground-truth query on the evaluation database. This metric, which directly measures end-task success, is standard in Spider and other NL2SQL evaluations. The SQL refinement loop for query generation is limited to $R=3$ attempts, balancing accuracy gains with inference cost.

\subsection{Qualitative Results}

We first present three success cases and two failure cases. Figures~\ref{fig:success_case_1}--\ref{fig:success_case_3} show instances where the baseline fails but our method succeeds, handling challenging scenarios such as multi-join queries, nested aggregations, and schema alignment. In contrast, Figures~\ref{fig:failure_case_1}--\ref{fig:failure_case_2} illustrate two failure cases of our method: errors in sub-question decomposition or the final CS lead to incorrect outputs.

\subsection{Quantitative Results}

\paragraph{Impact of Column Selection and Merging Strategy.} Table~\ref{table:impact_of_column_selection_and_merging_strategy} evaluates the effect of the CS refinement and the two merging strategies. Incorporating the CS step consistently improves performance across model sizes and strategies, typically by 2-5\% EX. For example, with 7B models using the Last Sub-query merge, adding the CS step raises accuracy from 72.26\% to 74.44\%. The Planner\&Executor strategy benefits especially from CS: without CS it sometimes underperforms the simpler (Last Sub-query) strategy (e.g., 66.77\% vs 72.26\% at 7B), but with CS it surpasses it (75.85\% vs 74.44\% at 7B). This can be attributed to the planner occasionally introducing extraneous or misordered columns that the final column-selection step fixes. While Planner\&Executor with CS yields the highest accuracy for each model size, it also incurs more latency than the Last Sub-query heuristic, highlighting a trade-off between accuracy and efficiency.

\paragraph{Divide-and-Merge Module and \textsc{AgentiQL}.} Table~\ref{table:divide_and_merge_module_performance} compares our divide-and-merge module (with CS) against the baseline across different model scales. With the integration of an effective routing mechanism, performance approaches that of the state of the art (SOTA), even when smaller reasoning and coding models compared to GPT-4o are employed. We also found that the relative performance of \textsc{AgentiQL} vs. the baseline correlates with database complexity: for larger LLMs, our pipeline tends to have a greater advantage on queries from schemas with many tables, whereas with a small model, the baseline performed better on the most complex schemas.


\section{Conclusion}


\textsc{AgentiQL} demonstrates that dividing complex natural language queries into sub-questions and merging their answers can improve interpretability while maintaining high accuracy in text-to-SQL generation. The CS refinement consistently yields additional gains in execution accuracy, with the Planner\&Executor merging strategy performing best when refinement is applied. Using an adaptive router to combine our pipeline with a strong baseline further enhances robustness and overall performance by exploiting their complementary strengths. Qualitative analyses show strong improvements in handling joins, nested aggregations, and schema alignment, though errors in decomposition and column refinement remain as failure cases. To facilitate reproducibility, we will release our code and prompt templates publicly.

\label{limitation_and_future_works}

Several limitations and directions remain for future work. We evaluated primarily on the Spider dataset; testing \textsc{AgentiQL} on additional benchmarks (e.g., BIRD or SQL-Eval) will be important to assess the generalizability of the findings. Experiments were limited to open-source LLMs up to 32B parameters, while larger models (and closed-source systems such as GPT-4 or Claude) were not fully explored due to resource constraints. Scaling to very large models incurs substantial cost: for example, in a preliminary trial, using a 235B-parameter model in our pipeline took nearly 60 minutes per question on four A100 GPUs (with CPU offloading).
There are multiple avenues for future research to address the above limitations. For instance, RL could be incorporated into the query generation stage, similar to what has been demonstrated in SkyRL-SQL~\cite{liu2025skyrlsql}, to provide adaptive feedback and further enhance the quality of generated SQL queries. Testing should also be extended to additional datasets such as BIRD and SQL-Eval, in order to evaluate the robustness of the approach across different benchmarks. Furthermore, experiments with closed-source LLMs, such as GPT-4o \cite{openai2024gpt4ocard} and Claude 4 \cite{claude4}, as well as larger open-source models, such as Qwen3-235B-A22B-Instruct and Qwen3-Coder-480B-A35B-Instruct, will provide deeper insights into the framework's scalability and general applicability. The effect of different merging strategies on accuracy and efficiency should be investigated, as should various routing options, such as XGBoost classifier and retrieval-augmented generation \cite{lewis2021retrievalaugmentedgenerationknowledgeintensivenlp}, to enable the framework to adapt more effectively to diverse question complexities and schema structures.

\bibliographystyle{unsrtnat}
\bibliography{bibliography}


\appendix

\section{Technical Appendices and Supplementary Material}
\begin{figure}[H]
    \centering
    \includegraphics[width=\linewidth]{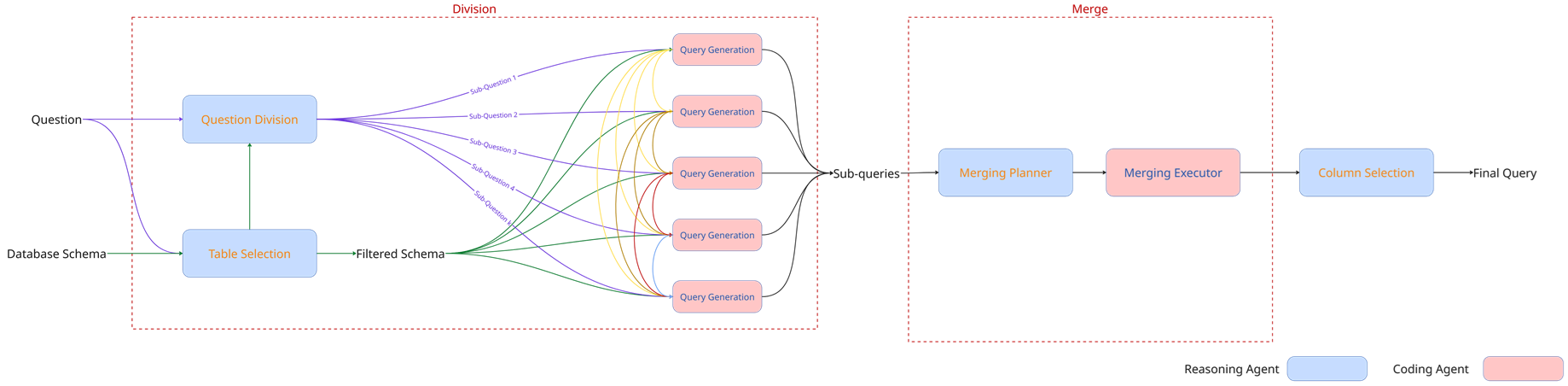}
    \caption{The Divide-and-Merge module of \textsc{AgentiQL}. The reasoning agent splits an input natural language query into multiple sub-questions. The coding agent then generates a corresponding SQL sub-query for each sub-question, and finally all sub-queries are merged to produce the final SQL query. This multi-step approach explicitly exposes intermediate reasoning steps and ensures the final query aligns with the question’s intent.}
    \label{fig:divide_and_merge_architecture}
\end{figure}

\begin{figure}[H]
    \centering
    \includegraphics[width=\linewidth]{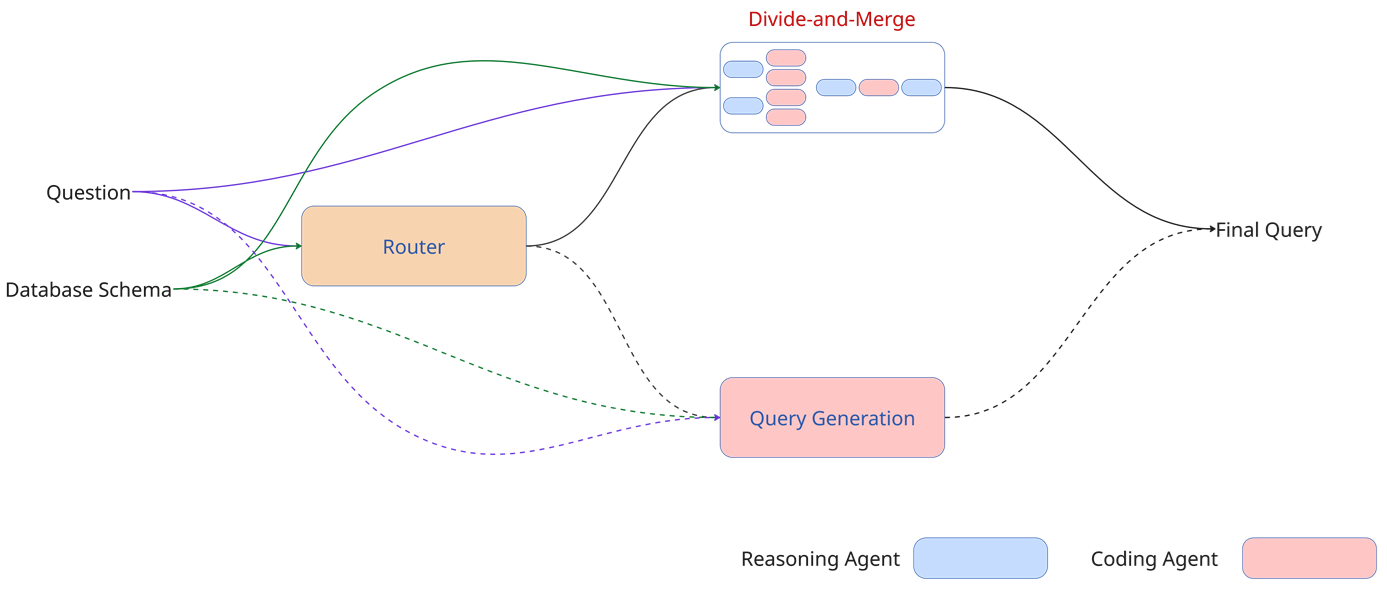}
    \caption{Overall architecture of \textsc{AgentiQL}. An input query is first evaluated by an adaptive router, which decides whether to send it to a one-step baseline parser or to the divide-and-merge module. The baseline directly generates the SQL query, while the divide-and-merge module processes the query through a multi-expert pipeline before producing the final SQL. This design balances interpretability, provided by the modular pipeline, with efficiency, provided by direct execution for simpler cases.}
    \label{fig:agentiQL_architecture}
\end{figure}

\begin{figure}[H]
\centering
\begin{tikzpicture}[node distance=7mm]
  \node[casebox, text width=\linewidth] (q)
    {\textbf{Question:} List the id, first name and last name of the customers who both have placed more than 2 orders and have bought at least 3 items.};

  \node[casebox, below=0.2cm of q, text width=\linewidth] (gold)
    {
    \textbf{Database Schema:}\\[2pt]
    \small
    \begin{itemize}
      \setlength\itemsep{1pt}      
      \setlength\leftskip{-15pt}     
    \item Products (product\_id, parent\_product\_id, product\_name, product\_price, product\_color, product\_size, product\_description)
    \item Customers (customer\_id, gender\_code, customer\_first\_name, 
    customer\_middle\_initial, customer\_last\_name, email\_address, login\_name, login\_password, phone\_number, address\_line\_1, town\_city, county, country)    
    \item Customer\_Payment\_Methods (customer\_id, payment\_method\_code)
    \item Invoices (invoice\_number, invoice\_status\_code, invoice\_date)
    \item Orders (order\_id, customer\_id, order\_status\_code, date\_order\_placed)
    \item Order\_Items (order\_item\_id, product\_id, order\_id, order\_item\_status\_code)
    \item Shipments (shipment\_id, order\_id, invoice\_number, shipment\_tracking\_number, shipment\_date)
    \item Shipment\_Items (shipment\_id, order\_item\_id)
    \end{itemize}
    };
  \node[casebox, below=0.15cm of gold, text width=\linewidth] (subqs)
    {\textit{Sub-question 1:} Find customers who have placed more than 2 orders.
    \newline \textit{Sub-question 2:} Join this result with the Orders table to get order details.
    \newline \textit{Sub-question 3:} Count the number of orders per customer.
    \newline \textit{Sub-question 4:} Filter to customers who have placed more than 2 orders.
    \newline \textit{Sub-question 5:} Find customers who have bought at least 3 items.
    \newline \textit{Sub-question 6:} Join this result with the Order\_Items table to get item details.
    \newline \textit{Sub-question 7:} Count the number of items per customer.
    \newline \textit{Sub-question 8:} Filter to customers who have bought at least 3 items.
    \newline \textit{Sub-question 9:} Find the intersection of customers from steps 4 and 8.
    \newline \textit{Sub-question 10:} Retrieve the id, first name, and last name of these customers.
    };
  \node[casebox, below=0.15cm of subqs, text width=\linewidth] (ours)
    {\textbf{Divide-and-Merge SQL:}\\[2pt]
     \begin{lstlisting}[style=sql]
SELECT customer_id, customer_first_name, customer_last_name FROM Customers
WHERE customer_id IN (
    SELECT c.customer_id FROM Customers c JOIN Orders o ON c.customer_id = o.customer_id GROUP BY c.customer_id HAVING COUNT(o.order_id) > 2
    INTERSECT
    SELECT c.customer_id FROM Customers c JOIN Orders o ON c.customer_id = o.customer_id JOIN Order_Items oi ON o.order_id = oi.order_id GROUP BY c.customer_id HAVING COUNT(oi.order_item_id) >= 3
);
  \end{lstlisting}};
  \node[casebox, below=0.2cm of ours, text width=\linewidth] (base)
    {\textbf{Baseline SQL:}\\[2pt]
     \begin{lstlisting}[style=sql]
SELECT c.customer_id, c.customer_first_name, c.customer_last_name FROM Customers c JOIN Orders o ON c.customer_id = o.customer_id GROUP BY c.customer_id HAVING COUNT(o.order_id) > 2
INTERSECT
SELECT c.customer_id, c.customer_first_name, c.customer_last_name FROM Customers c JOIN Order_Items oi ON c.customer_id = oi.customer_id GROUP BY c.customer_id HAVING COUNT(oi.order_item_id) >= 3;
  \end{lstlisting}};
  \node[fit=(base), draw=red!70, rounded corners, thick, inner sep=2pt, label={[red!70!black]below:{\footnotesize Invalid join path: incorrect linking of Customers directly to Order\_Items}}] {};
\end{tikzpicture}
\caption{Success case with Qwen2.5-7B-Instruct and Qwen2.5-Coder-7B-Instruct using the Planner\&Executor merging strategy. The query requires finding customers with both more than two orders and at least three items. The baseline SQL fails by joining \texttt{Customers} directly with \texttt{Order\_Items}, violating the schema. The Divide-and-Merge approach, however, decomposes the task, routes through \texttt{Orders}, and intersects constraints, producing a valid SQL that retrieves the correct customers.}
\label{fig:success_case_1}
\end{figure}

\begin{figure}[H]
\centering
\begin{tikzpicture}[node distance=7mm]
  \node[casebox, text width=\linewidth] (q)
    {\textbf{Question:} What is the price of all products being ordered on average?};

  \node[casebox, below=0.2cm of q, text width=\linewidth] (gold)
    {
    \textbf{Database Schema:}\\[2pt]
    \small
    \begin{itemize}
      \setlength\itemsep{1pt}      
      \setlength\leftskip{-15pt}     
    \item Products (product\_id, parent\_product\_id, product\_name, product\_price, product\_color, product\_size, product\_description)
    \item Customers (customer\_id, gender\_code, customer\_first\_name, 
    customer\_middle\_initial, customer\_last\_name, email\_address, login\_name, login\_password, phone\_number, address\_line\_1, town\_city, county, country)    
    \item Customer\_Payment\_Methods (customer\_id, payment\_method\_code)
    \item Invoices (invoice\_number, invoice\_status\_code, invoice\_date)
    \item Orders (order\_id, customer\_id, order\_status\_code, date\_order\_placed)
    \item Order\_Items (order\_item\_id, product\_id, order\_id, order\_item\_status\_code)
    \item Shipments (shipment\_id, order\_id, invoice\_number, shipment\_tracking\_number, shipment\_date)
    \item Shipment\_Items (shipment\_id, order\_item\_id)
    \end{itemize}
    };
  \node[casebox, below=0.15cm of gold, text width=\linewidth] (subqs)
    {\textit{Sub-question 1:} Find the price of each product.
    \newline \textit{Sub-question 2:} Join the Orders and Order\_Items tables to associate orders with their items.
    \newline \textit{Sub-question 3:} Join the resulting dataset with the Products table to get the prices of the ordered products.
    \newline \textit{Sub-question 4:} Calculate the average price of the ordered products.
    };
  \node[casebox, below=0.15cm of subqs, text width=\linewidth] (ours)
    {\textbf{Divide-and-Merge SQL:}\\[2pt]
     \begin{lstlisting}[style=sql]
SELECT AVG(p.product_price) AS average_price
FROM Order_Items oi
JOIN Products p ON oi.product_id = p.product_id;
  \end{lstlisting}};
  \node[casebox, below=0.2cm of ours, text width=\linewidth] (base)
    {\textbf{Baseline SQL:}\\[2pt]
     \begin{lstlisting}[style=sql]
SELECT AVG(product_price) 
FROM Products;
  \end{lstlisting}};
  \node[fit=(base), draw=red!70, rounded corners, thick, inner sep=2pt, label={[red!70!black]below:{\footnotesize Ignores orders: averages all products (missing join to Order\_Items)}}] {};
\end{tikzpicture}
\caption{Success case with Qwen2.5-14B-Instruct and Qwen2.5-Coder-14B-Instruct using the Last-Subquery merging strategy. The query asks for the average price of products that have been ordered. The baseline SQL incorrectly averages all products in \texttt{Products}, ignoring order information. In contrast, the Divide-and-Merge approach decomposes the task, joins \texttt{Order\_Items} with \texttt{Products}, and computes the average over ordered products only, yielding the correct result.}

\label{fig:success_case_2}
\end{figure}

\begin{figure}[H]
\centering
\begin{tikzpicture}[node distance=7mm]
  \node[casebox, text width=\linewidth] (q)
    {\textbf{Question:} Please show the most common affiliation for city channels.};

  \node[casebox, below=0.2cm of q, text width=\linewidth] (gold)
    {
    \textbf{Database Schema:}\\[2pt]
    \small
    \begin{itemize}
      \setlength\itemsep{1pt}      
      \setlength\leftskip{-15pt}     
  \item city\_channel (ID, City, Station\_name, Owned\_Since, Affiliation)
  \item radio (Radio\_ID, Transmitter, Radio\_MHz, 2FM\_MHz, RnaG\_MHz, Lyric\_FM\_MHz, ERP\_kW)
  \item tv\_show (tv\_show\_ID, tv\_show\_name, Sub\_tittle, Next\_show\_name, Original\_Airdate)
  \item city\_channel\_radio (City\_channel\_ID, Radio\_ID, Is\_online)
  \item city\_channel\_tv\_show (City\_channel\_ID, tv\_show\_ID, Is\_online, Is\_free)
    \end{itemize}
    };
  \node[casebox, below=0.15cm of gold, text width=\linewidth] (subqs)
    {\textit{Sub-question 1:} Select the Affiliation column from the city\_channel table.
    \newline \textit{Sub-question 2:} Select the Affiliation column from the city\_channel table.
    \newline \textit{Sub-question 3:} Group the results by Affiliation.
    \newline \textit{Sub-question 4:} Count the number of occurrences for each Affiliation.
    \newline \textit{Sub-question 5:}  Order the groups by the count in descending order.
    \newline \textit{Sub-question 6:} Select the top group's Affiliation.
    };
  \node[casebox, below=0.15cm of subqs, text width=\linewidth] (ours)
    {\textbf{Divide-and-Merge SQL:}\\[2pt]
     \begin{lstlisting}[style=sql]
SELECT Affiliation
FROM (
    SELECT Affiliation, COUNT(*) AS count
    FROM city_channel
    GROUP BY Affiliation
    ORDER BY count DESC
) AS grouped_affiliations
LIMIT 1;
  \end{lstlisting}};
  \node[casebox, below=0.2cm of ours, text width=\linewidth] (base)
    {\textbf{Baseline SQL:}\\[2pt]
     \begin{lstlisting}[style=sql]
SELECT Affiliation, COUNT(*) AS count
FROM city_channel
GROUP BY Affiliation
ORDER BY count DESC
LIMIT 1;
  \end{lstlisting}};
  \node[fit=(base), draw=red!70, rounded corners, thick, inner sep=2pt, label={[red!70!black]below:{\footnotesize Extra output: baseline returns counts instead of only the affiliation}}] {};
\end{tikzpicture}
\caption{Success case with Qwen2.5-32B-Instruct and Qwen2.5-Coder-32B-Instruct using the Planner\&Executor merging strategy. The query asks for the most common affiliation among city channels. The baseline SQL outputs both the affiliation and its count, adding extra information. The Divide-and-Merge approach refines the output to return only the affiliation, exactly matching the query intent.}

\label{fig:success_case_3}
\end{figure}

\begin{figure}[H]
\centering
\begin{tikzpicture}[node distance=7mm]
  \node[casebox, text width=\linewidth] (q)
    {\textbf{Question:} Show agency ids and the number of clients for each agency.};

  \node[casebox, below=0.2cm of q, text width=\linewidth] (gold)
    {
    \textbf{Database Schema:}\\[2pt]
    \small
    \begin{itemize}
      \setlength\itemsep{1pt}      
      \setlength\leftskip{-15pt}     
  \item Agencies (agency\_id, agency\_details)
  \item Staff (staff\_id, agency\_id, staff\_details)
  \item Clients (client\_id, agency\_id, sic\_code, client\_details)
  \item Invoices (invoice\_id, client\_id, invoice\_status, invoice\_details)
  \item Meetings (meeting\_id, client\_id, meeting\_outcome, meeting\_type, billable\_yn, start\_date\_time, end\_date\_time, purpose\_of\_meeting, other\_details)
  \item Payments (payment\_id, invoice\_id, payment\_details)
  \item Staff\_in\_Meetings (meeting\_id, staff\_id)
    \end{itemize}
    };
  \node[casebox, below=0.15cm of gold, text width=\linewidth] (subqs)
    {\textit{Sub-question 1:} Join the Agencies table with the Clients table on agency\_id.
    \newline \textit{Sub-question 2:} Count the number of clients for each agency.
    \newline \textit{Sub-question 3:} Select the agency\_id and the count of clients.
    };
  \node[casebox, below=0.15cm of subqs, text width=\linewidth] (ours)
    {\textbf{Divide-and-Merge SQL:}\\[2pt]
     \begin{lstlisting}[style=sql]
SELECT agency_id, COUNT(client_id) AS client_count
FROM Agencies a
JOIN Clients c ON a.agency_id = c.agency_id
GROUP BY agency_id;
  \end{lstlisting}};
  \node[fit=(ours), draw=red!70, rounded corners, thick, inner sep=2pt, label={[red!70!black]below:{\footnotesize Ambiguous column: agency\_id not qualified after join }}] {};
  \node[casebox, below=0.7cm of ours, text width=\linewidth] (base)
    {\textbf{Baseline SQL:}\\[2pt]
     \begin{lstlisting}[style=sql]
SELECT agency_id, COUNT(client_id) AS client_count
FROM Clients
GROUP BY agency_id;
  \end{lstlisting}};
\end{tikzpicture}
\caption{Failure case with Qwen2.5-7B-Instruct and Qwen2.5-Coder-7B-Instruct using the Planner\&Executor merging strategy. The query asks for agency IDs with their client counts. The Divide-and-Merge SQL introduces an explicit join between \texttt{Agencies} and \texttt{Clients}, but fails due to an ambiguous reference to \texttt{agency\_id}. The baseline SQL, grouping directly over \texttt{Clients}, avoids this error and produces the correct result with a simpler query.}

\label{fig:failure_case_1}
\end{figure}

\begin{figure}[H]
\centering
\begin{tikzpicture}[node distance=7mm]
  \node[casebox, text width=\linewidth] (q)
    {\textbf{Question:} Return the number of stories for each building in the region named ``Abruzzo''.};

  \node[casebox, below=0.2cm of q, text width=\linewidth] (gold)
    {
    \textbf{Database Schema:}\\[2pt]
    \small
    \begin{itemize}
      \setlength\itemsep{1pt}      
      \setlength\leftskip{-15pt}     
  \item building (Building\_ID, Region\_ID, Name, Address, Number\_of\_Stories, Completed\_Year)
  \item region (Region\_ID, Name, Capital, Area, Population)
    \end{itemize}
    };
  \node[casebox, below=0.15cm of gold, text width=\linewidth] (subqs)
    {\textit{Sub-question 1:} Find the Region\_ID for the region named "Abruzzo".
    \newline \textit{Sub-question 2:} Filter buildings that belong to the region with the found Region\_ID.
    \newline \textit{Sub-question 3:} Retrieve the Name and Number\_of\_Stories for these buildings.
    };
  \node[casebox, below=0.15cm of subqs, text width=\linewidth] (ours)
    {\textbf{Divide-and-Merge SQL:}\\[2pt]
     \begin{lstlisting}[style=sql]
SELECT b.Name, b.Number_of_Stories
FROM building b
JOIN (
    SELECT Region_ID
    FROM region
    WHERE Name = 'Abruzzo'
) AS r ON b.Region_ID = r.Region_ID;
  \end{lstlisting}};
  \node[fit=(ours), draw=red!70, rounded corners, thick, inner sep=2pt, label={[red!70!black]below:{\footnotesize Extra output: divide-and-merge returns building names in addition to number of stories }}] {};
  \node[casebox, below=0.7cm of ours, text width=\linewidth] (base)
    {\textbf{Baseline SQL:}\\[2pt]
     \begin{lstlisting}[style=sql]
SELECT b.Number_of_Stories
FROM building b
JOIN region r ON b.Region_ID = r.Region_ID
WHERE r.Name = 'Abruzzo';
  \end{lstlisting}};
\end{tikzpicture}
\caption{Failure case with Qwen2.5-32B-Instruct and Qwen2.5-Coder-32B-Instruct using the Last-Sub-query merging strategy. The query asks for the number of stories for each building in the region ``Abruzzo.'' The Divide-and-Merge SQL correctly filters buildings by region but incorrectly adds building names to the output, returning more information than required. The baseline SQL directly joins \texttt{building} and \texttt{region} and outputs only \texttt{Number\_of\_Stories}, exactly matching the query intent.}
\label{fig:failure_case_2}
\end{figure}

\begin{table}[H]
\centering
\begin{tabular}{|c c c c c|}
\hline
Reasoning Agent & Coding Agent & Merging Strategy & w/o CS & with CS\\ [0.5ex]
\hline
\multirow{2}{*}{Qwen2.5-7B-Instruct} & \multirow{2}{*}{Qwen2.5-Coder-7B-Instruct} & Last Sub-query & 72.26 & 74.44 \\
& & Planner\&Executor & 66.77 & \textbf{75.85} \\ \hline
\multirow{2}{*}{Qwen2.5-14B-Instruct} & \multirow{2}{*}{Qwen2.5-Coder-14B-Instruct} & Last Sub-query & 69.33 & 76.61 \\ 
 &  & Planner\&Executor & 74.27 & \textbf{77.16} \\ \hline
\multirow{2}{*}{Qwen2.5-32B-Instruct} & \multirow{2}{*}{Qwen2.5-Coder-32B-Instruct} & Last Sub-query & 73.84 & 78.57 \\
 & & Planner\&Executor & 75.58 & \textbf{79.77} \\ \hline
\end{tabular}
\caption{Impact of CS refinement on EX(\%) for the Spider test set. Results are shown for different combinations among reasoning agent, coding agent, and merging strategy. Incorporating CS consistently improves performance across settings, with gains of up to 9\% compared to models without refinement. The Planner\&Executor strategy benefits the most from CS, showing that finer control over column choices enhances alignment with user intent.}
\label{table:impact_of_column_selection_and_merging_strategy}
\end{table}

\begin{table}[H]
\centering
\resizebox{\textwidth}{!}{
\begin{tabular}{|c c c c c|}
\hline
Reasoning Agent & Coding Agent & Merging Strategy & \textsc{AgentiQL} Only & Baseline Only\\ [0.5ex]
\hline
\multirow{2}{*}{Qwen2.5-7B-Instruct} & \multirow{2}{*}{Qwen2.5-Coder-7B-Instruct} & Last Sub-query & 6.96 & 8.91 \\
 & & Planner\&Executor & 7.34 & 8.21 \\ \hline
 \multirow{2}{*}{Qwen2.5-14B-Instruct} & \multirow{2}{*}{Qwen2.5-Coder-14B-Instruct} & Last Sub-query & 5.00 & 9.10 \\
 & & Planner\&Executor & 4.78 & 8.91 \\ \hline
 \multirow{2}{*}{Qwen2.5-32B-Instruct} & \multirow{2}{*}{Qwen2.5-Coder-32B-Instruct} & Last Sub-query & 4.94 &  6.30 \\
 & & Planner\&Executor & 4.78 & 5.54 \\ \hline
\end{tabular}
}

\caption{Comparison of instances where \textsc{AgentiQL} and the baseline model differ in EX on the Spider test set. The columns report proportion of instances where our method succeeds but the baseline fails (\textbf{\textsc{AgentiQL} Only}) and cases where the baseline succeeds but ours fails (\textbf{Baseline Only}). Results show that while both methods capture distinct strengths, the Planner\&Executor merging strategy reduces the gap relative to the baseline.}
\label{table:percetage_of_wrong_and_correct_among_ours_and_baseline}
\end{table}

\begin{table}[H]
\centering
\resizebox{\textwidth}{!}{
\begin{tabular}{|c c c c c c|}
\hline
Method & Reasoning Agent & Coding Agent & Merging Strategy & Spider-Test & \\ [0.5ex]
\hline
Baseline & \multirow{3}{*}{Qwen2.5-7B-Instruct} &  \multirow{3}{*}{Qwen2.5-Coder-7B-Instruct}  & - & 76.4 & \\ 
\multirow{2}{*}{Ours} & & & Last Sub-query & 74.44 & \\ 
 &  &  & Planner\&Executor & 75.85 & \\ \hline

Baseline & \multirow{3}{*}{Qwen2.5-14B-Instruct} &  \multirow{3}{*}{Qwen2.5-Coder-14B-Instruct}  & - & 80.80 & \\ 
\multirow{2}{*}{Ours} &  &  &Last Sub-query & 76.61 & \\ 
 &  &  & Planner\&Executor & 77.16 & \\ \hline

Baseline & \multirow{3}{*}{Qwen2.5-32B-Instruct} &  \multirow{3}{*}{Qwen2.5-Coder-32B-Instruct} & - & 79.93 & \\ 
\multirow{2}{*}{Ours} & & & Last Sub-query & 78.57 & \\ 
 &  &  & Planner\&Executor & 79.77 & \\ \hline

XiYan-SQL \cite{gao2025previewxiyansqlmultigeneratorensemble} & GPT-4o & GPT-4o & - & 89.65 & \\ \hline
\end{tabular}
}
\caption{EX(\%) on the Spider test set for the Divide-and-Merge module. Results are reported for Qwen2.5 models of varying sizes. While the baseline achieves strong performance, the Planner\&Executor merging strategy improves over the naive last-sub-query approach, demonstrating the benefit of decomposition. Larger models generally yield higher accuracy.}
\label{table:divide_and_merge_module_performance}
\end{table}

\begin{table}[H]
\centering
\resizebox{\textwidth}{!}{
\begin{tabular}{|c c c c c|}
\hline
Reasoning Agent & Coding Agent & Merging Strategy & Pearson & Spearmanr \\ [0.5ex]
\hline
\multirow{2}{*}{Qwen2.5-7B-Instruct} & \multirow{2}{*}{Qwen2.5-Coder-7B-Instruct} & Last Sub-query & -0.61 & -0.74 \\
 & & Planner\&Executor & -0.39 & -0.79 \\ \hline
 \multirow{2}{*}{Qwen2.5-14B-Instruct} & \multirow{2}{*}{Qwen2.5-Coder-14B-Instruct} & Last Sub-query & 0.46 & 0.70 \\
 & & Planner\&Executor & 0.37 & 0.52 \\ \hline
 \multirow{2}{*}{Qwen2.5-32B-Instruct} & \multirow{2}{*}{Qwen2.5-Coder-32B-Instruct} & Last Sub-query & 0.35 &  0.67 \\
 & & Planner\&Executor & 0.12 & 0.64 \\ \hline
\end{tabular}
}
\caption{Correlation between schema complexity and performance improvements of our method over the baseline. Schema complexity is measured using a simple metric, which is computed by calculation of the number of tables in the schema. Pearson \cite{pearson1895} and Spearman \cite{spearman1904} coefficients are reported for different combinations of reasoning agents, coding agents, and merging strategies.}
\label{table:corr_table_count_vs_performance}
\end{table}

\begin{figure}[H]
    \centering
    \includegraphics[width=0.8\linewidth]{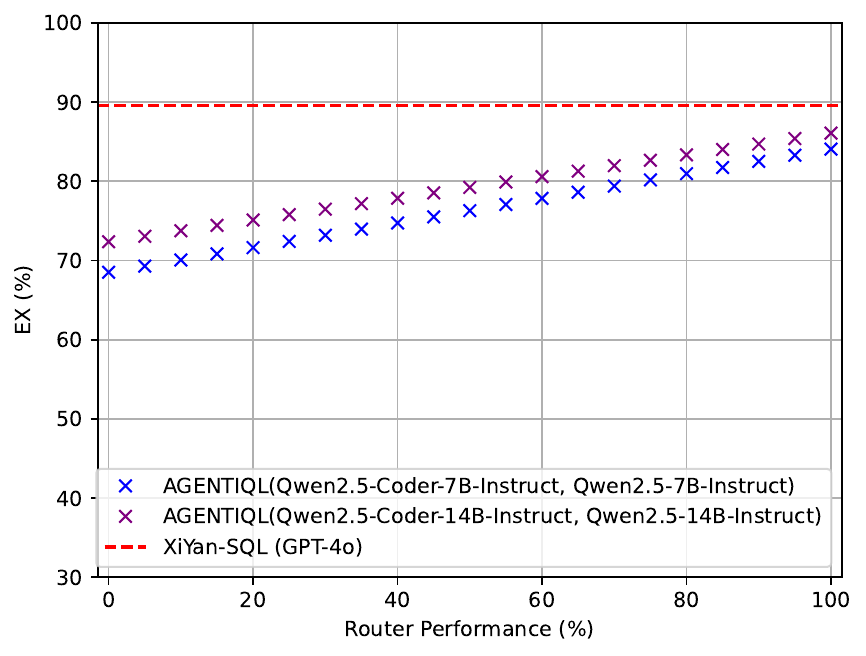}
    \caption{EX(\%) on the Spider test set plotted against the adaptive router’s accuracy. The curves show AGENTIQL’s performance under two configurations (with different model sizes) using the Planner\&Executor merging strategy compared to the state-of-the-art system XiYan-SQL. \textsc{AgentiQL} employs dynamic routing that selects either a modular pipeline (divide-and-merge approach) or a baseline parser for each query. As the router’s accuracy increases (i.e., more queries are routed correctly), \textsc{AgentiQL}’s execution accuracy steadily improves, approaching the level of the SOTA system despite using much smaller models. This highlights the effectiveness of dynamic routing in narrowing the performance gap to state-of-the-art solutions while maintaining efficiency.}
    \label{fig:agentiQL_performance_sota_compare}
\end{figure}

\section{Experimental Setup}
\label{experimental_setup}
All local experiments were conducted on an internal compute cluster equipped with NVIDIA A100 GPUs each with 80GB of memory. A total of eight GPUs were available, as confirmed by system diagnostics. For open-source models, we estimate GPU memory requirements based on parameter size: Qwen2.5-7B-Instruct(7B parameters), Qwen2.5-14B-Instruct(14B parameters), Qwen2.5-32B-Instruct (32B parameters), Qwen2.5-Coder-7B-Instruct (7B parameters), Qwen2.5-Coder-14B-Instruct (14B parameters), and Qwen2.5-Coder-32B-Instruct (32B parameters).
Some experiments were executed in parallel because they had no dependencies, while others were computed sequentially due to dependency requirements. The total compute estimate for open-source models amounts to approximately 1450 GPU-hours.

\section{Assets License}
\label{assets_license}
We evaluated the following LLMs on the Spider \cite{yu2019spiderlargescalehumanlabeleddataset} dataset. Below, we list each asset along with its creator and the corresponding license or usage information, where available. \\
\textbf{LLMs}
\begin{itemize}
    \item Qwen2.5-7B-Instruct
    \begin{itemize}
      \item Creator: Alibaba Cloud
      \item License: Apache license 2.0, \href{https://huggingface.co/Qwen/Qwen2.5-7B-Instruct}{Hugging Face link}
    \end{itemize}
    \item Qwen2.5-14B-Instruct
    \begin{itemize}
      \item Creator: Alibaba Cloud
      \item License: Apache license 2.0, \href{https://huggingface.co/Qwen/Qwen2.5-14B-Instruct}{Hugging Face link}
    \end{itemize}
    \item Qwen2.5-32B-Instruct
    \begin{itemize}
      \item Creator: Alibaba Cloud
      \item License: Apache license 2.0, \href{https://huggingface.co/Qwen/Qwen2.5-32B-Instruct}{Hugging Face link}
    \end{itemize}
    \item Qwen2.5-Coder-7B-Instruct
    \begin{itemize}
      \item Creator: Alibaba Cloud
      \item License: Apache license 2.0, \href{https://huggingface.co/Qwen/Qwen2.5-Coder-7B-Instruct}{Hugging Face link}
    \end{itemize}
    \item Qwen2.5-Coder-14B-Instruct
    \begin{itemize}
      \item Creator: Alibaba Cloud
      \item License: Apache license 2.0, \href{https://huggingface.co/Qwen/Qwen2.5-Coder-14B-Instruct}{Hugging Face link}
    \end{itemize}
    \item Qwen2.5-Coder-32B-Instruct
    \begin{itemize}
      \item Creator: Alibaba Cloud
      \item License: Apache license 2.0, \href{https://huggingface.co/Qwen/Qwen2.5-Coder-32B-Instruct}{Hugging Face link}
    \end{itemize}
    \item GPT-4o
    \begin{itemize}
      \item Creator: OpenAI
      \item License: Accessed via API under \href{https://openai.com/policies/}{OpenAI Terms of Use}
    \end{itemize}
\end{itemize}

\end{document}